\documentclass[a4paper]{cas-sc}

\usepackage[numbers,sort&compress]{natbib}
\usepackage{graphicx}%
\usepackage{multirow}%
\usepackage{amsmath,amssymb,amsfonts}%
\usepackage{amsthm}%
\usepackage{mathrsfs}%
\usepackage[title]{appendix}%
\usepackage{xcolor}%
\usepackage{textcomp}%
\usepackage{manyfoot}%
\usepackage{booktabs}%
\usepackage{amsmath}
\usepackage{algorithm}
\usepackage{algpseudocode}
\usepackage{amsmath}
\usepackage{listings}%
\usepackage{hyperref} 
\usepackage{tabularx}
\usepackage{array}

\def\tsc#1{\csdef{#1}{\textsc{\lowercase{#1}}\xspace}}
\tsc{WGM}
\tsc{QE}

\begin{document}
\let\WriteBookmarks\relax
\def\floatpagepagefraction{1}
\def\textpagefraction{.001}

\shorttitle{}    

\shortauthors{}  

\title [mode = title]{PrototypeFormer: Learning to Explore Prototype Relationships for  Few-shot Image Classification}  

\tnotemark[1] 

\tnotetext[1]{} 

%

\author[1]{Meijuan Su}





\credit{Conceptualization, Methodology, Software, Validation, Writing - review and editing}

\affiliation[1]{organization={School of Computer Science and Technology, Soochow University},
            postcode={215000}, 
            city={Suzhou},
            country={China}}

\author[2]{Feihong He}


\credit{Conceptualization, Methodology, Software, Writing - original draft, Writing - review and editing, Validation}

\affiliation[2]{organization={School of Cyberspace Security, Sun Yat-sen University},
            postcode={518107}, 
            city={Shenzhen},
            country={China}}

\author[1]{Fanzhang Li \corref{cor1}}
\credit{Investigation, Methodology, Software, Project administration, Writing - review. }




\cortext[cor1]{Corresponding author}




\begin{abstract}
Few-shot image classification has received considerable attention for overcoming the challenge of limited classification performance with limited samples in novel classes. Most existing works employ sophisticated learning strategies and feature learning modules to alleviate this challenge. In this paper, we propose a novel method called PrototypeFormer, exploring the relationships among category prototypes in the few-shot scenario. Specifically, we utilize a transformer architecture to build a prototype extraction module, aiming to extract class representations that are more discriminative for few-shot classification. Besides, during the model training process, we propose a contrastive learning-based optimization approach to optimize prototype features in few-shot learning scenarios. Despite its simplicity, our method performs remarkably well, with no bells and whistles. We have experimented with our approach on several popular few-shot image classification benchmark datasets, which shows that our method outperforms all current state-of-the-art methods. In particular, our method achieves 97.07\% and 90.88\% on 5-way 5-shot and 5-way 1-shot tasks of miniImageNet, which surpasses the state-of-the-art results with accuracy of 0.57\% and 6.84\%, respectively. The code will be released later.  
\end{abstract}


\begin{highlights}
\item \textbf{Prototype Extraction Module. }  We introduce a novel and efficient transformer-based architecture specifically designed for few-shot learning. This module, termed the Prototype Extraction Module, leverages the self-attention mechanism of transformers to capture intricate relationships among intra-class samples. By treating class prototypes as learnable tokens and integrating them with support set embeddings, the module extracts highly discriminative prototype representations. Unlike traditional methods that rely on global average pooling or local descriptors, our approach provides a comprehensive global perspective, enabling the model to better capture task-specific feature relationships. This module is simple yet powerful, significantly enhancing the model's ability to generalize in few-shot scenarios. 
\item \textbf{Prototype Contrastive Loss.} We form sub-prototypes by employing linear combinations of the support set. Subsequently, we optimize the model using the prototype contrastive loss based on these sub-prototypes to obtain more robust prototype representations. This approach ensures that similar class embeddings are pulled closer together, while dissimilar ones are pushed apart, leading to more robust and generalizable prototype representations. The contrastive loss is particularly effective in few-shot settings, where limited data makes traditional methods prone to overfitting.
\item \textbf{Achieving State-of-the-Art Performance.} We extensively evaluate our method on multiple widely used few-shot learning benchmarks. Our experiments demonstrate that PrototypeFormer consistently outperforms existing state-of-the-art methods across these datasets. Notably, on the miniImageNet dataset, our method achieves remarkable accuracy improvements of 0.57\% and 6.84\% for 5-way 5-shot and 5-way 1-shot tasks, respectively. These results highlight the effectiveness of our approach in addressing the challenges of few-shot learning, particularly in scenarios with limited labeled data. The success of our method is further validated by its strong performance on fine-grained classification tasks, such as those in the CUB-200 dataset.
\end{highlights}

\maketitle











\section{Introduction}

Neural networks have been remarkably successful in large-scale image classification. However, the domain of few-shot image classification, where models must rapidly adapt to new data distributions with limited labeled samples (e.g., five or one sample for each class), remains a challenge. As a result of its promising applications in diverse fields such as medical image analysis and robotics, few-shot learning ~\cite{survey} has captivated the attention of the computer vision and machine learning community.

Recent few-shot learning approaches mainly improve the generalization by augmenting the samples/features or facilitating feature representation with novel neural modules. A multitude of methods~\cite{metagan,mixup,MLS,DeltaEncoder,DC} utilizes generative models to generate new samples or augment feature space, aiming to approximate the actual distribution. Devising sophisticated feature representation modules is also a meaningful way to improve the model performance on low-shot categories. Specifically, CAN~\cite{CrossAttNet} leverages cross-attention mechanisms to acquire enriched sample embeddings with enhanced class-specific features in a transductive way, while DN4~\cite{DN4}, DMN4~\cite{DMN4}, and MCL~\cite{MCL} adopt local feature representations instead of global representations to obtain more discriminative feature representations. Following the line of feature representation learning approaches, we introduce a prototype extraction module to enhance the prototype embeddings. Contrary to earlier feature representation methodologies, our study delves into the intricate interconnections both within each class and across the entire task to derive more discriminative prototype representations.

\begin{figure}[!t]
\centering
\includegraphics[width=0.65\linewidth]{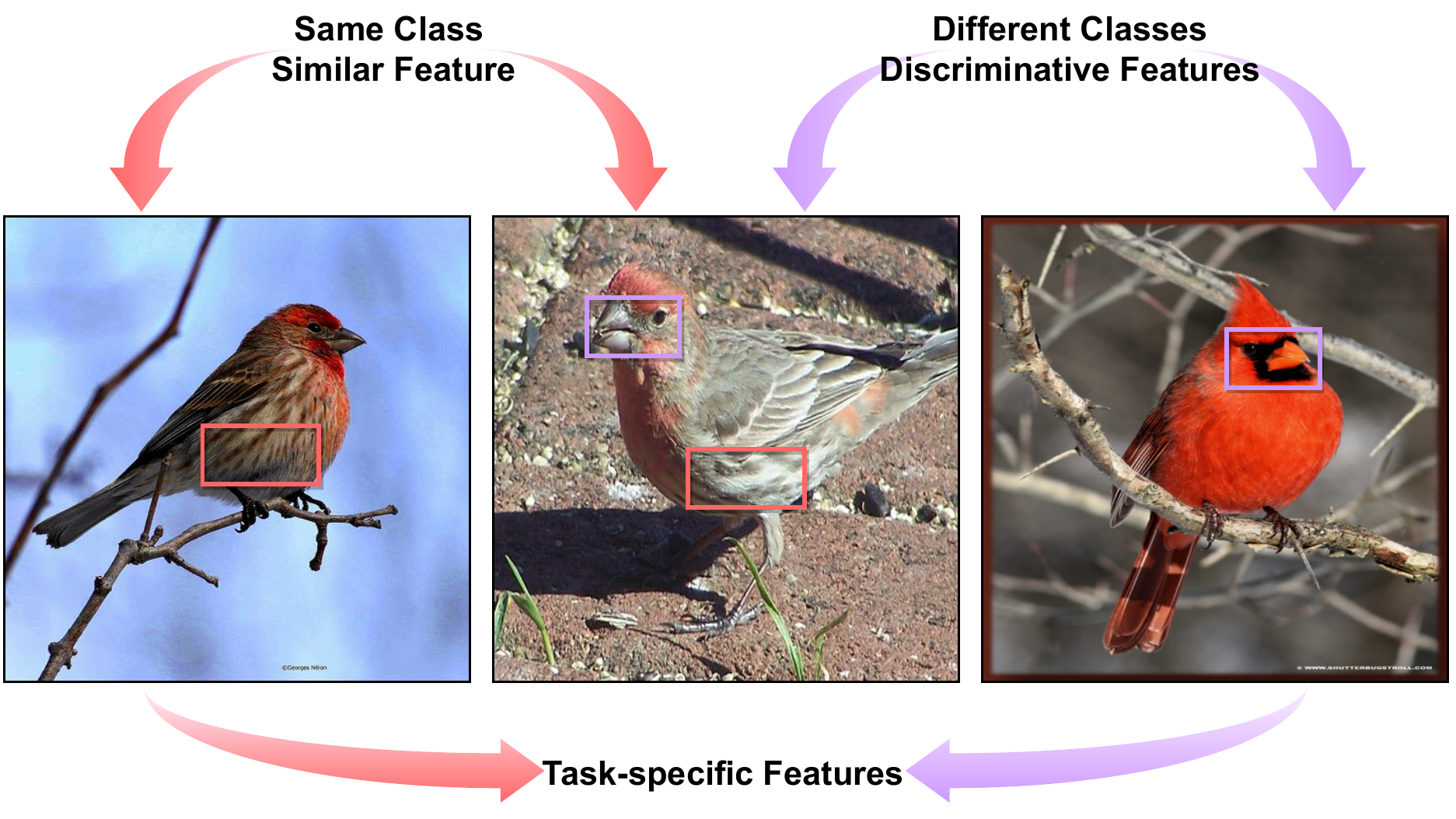}
\caption{Samples from different categories exhibit both shared features and distinctive features. For example, the red rectangle indicates the similarity features among different categories, while the purple rectangle represents dissimilar features across different categories.}
\label{headfigure}
\end{figure}

Learning prototype embedding~\cite{protonet,protocomplet} is useful for few-shot classification.
ProtoNet~\cite{protonet} introduces a methodology employing prototype points to encapsulate the feature embeddings of entire categories, and ~\cite{protocomplet} proposes to enhance the notion of prototype points. However, they significantly ignore the prototype relationships for learning robust class features. In this paper, we delve into the interconnections between prototype points, considering both intra-class and inter-class relationships. We first introduce a novel prototype extraction module to learn the relationship of intra-class samples through the self-attention of sub-prototypes. This module excels at obtaining a comprehensive global perspective, enabling the extraction of robust class features based on relationships among categories throughout the entire task. 

To further fortify the robustness of class features in few-shot scenarios, we introduce prototype contrastive loss, a novel contrastive loss designed explicitly to capture interactions among inter-class prototypes. One important concept in our approach is sub-prototypes, representing the average features of subsets of samples within each category. By employing these sub-prototypes within a contrastive learning framework, we aim to cultivate more discriminative representations. Specifically, the contrastive learning strategy ensures that similar class embeddings are drawn closer in the feature space, while dissimilar ones are pushed apart, thus enhancing the discriminative power of our representative prototypes.


Moreover, some works~\cite{CoOp,ClipAdapter} have demonstrated the impressive feature extraction capabilities of the CLIP pre-trained model in few-shot learning. 
As a result, we integrate CLIP into our approach, undertaking only a limited amount of parameter training. We conclude our contribution as follows:
\begin{itemize}
\item \textbf{Prototype Extraction Module. }  We introduce a novel and simple transformer-based architecture for few-shot learning, employing a learnable prototype extraction module to extract prototype representations.
\item \textbf{Prototype Contrastive Loss.} We form sub-prototypes by employing linear combinations of the support set. Subsequently, we optimize the model using the prototype contrastive loss based on these sub-prototypes to obtain more robust prototype representations. 
\item \textbf{Achieving State-of-the-Art Performance.} We evaluate our method on multiple publicly few-shot benchmark datasets, and the results demonstrate that the proposed method in this paper outperforms state-of-the-art few-shot learning methods across these datasets, achieving a remarkable improvement of up to 6.84\%. 
\end{itemize}

\begin{figure*}[!t]
\centering
\includegraphics[width=0.9\linewidth]{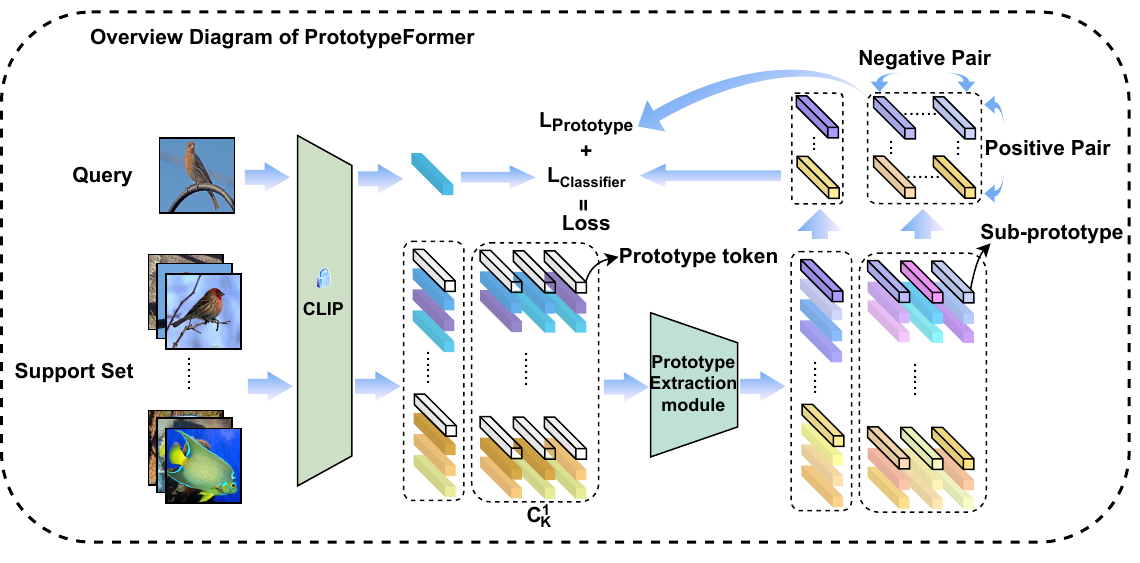}
\caption{This figure presents the overall process flowchart of the method proposed in this paper. We linearly combine the support set and obtain sub-prototypes through the prototype extraction module. The sub-prototypes are utilized for computing the prototype contrastive loss $L_{prototype}$, while the prototype is employed for calculating the classification loss $L_{classifier}$. We sum the $L_{prototype}$ and $L_{classifier}$ to obtain the final optimization objective.}
\label{overview}
\end{figure*}

\section{Related Work}
\subsection{Few-shot Learning}
The rapid development of deep neural networks in recent years has primarily benefited from large-scale labeled datasets. 
However, the high cost of data collection and manual labeling has brought few-shot learning to the forefront of widespread interest. Few-shot learning is usually classified into optimization-based and metric-based methods. 
The main idea of metric-based methods is to define specific metrics to classify samples in a way similar to the nearest neighbor algorithm.  The Siamese Network\cite{SiamNet} employs shared feature extractors to derive feature representations from both support sets and query sets. Subsequently, it computes classification similarity individually for each pair of support set and query set. Furthermore, the Siamese Network effectively distinguishes between different categories by comparing deep features of the support and query sets, maintaining high classification accuracy even when differences between categories are subtle. The Prototypical Network~\cite{protonet} computes prototype points for each class of samples, and the query samples are categorized by calculating the L2 distance to each prototype point. In Relation Network~\cite{RelatNet}, the incorporation of learnable nonlinear classifiers for sample classification is done innovatively. 
CAN~\cite{CrossAttNet} has improved model performance by computing cross-attention on samples to enhance the network's focus on classification targets. 
Also, to reduce sample background interference, local descriptors that do not contain classification targets are eliminated in DN4~\cite{DN4} and DMN4~\cite{DMN4} by comparing the similarity between local descriptors. 
COSOC~\cite{COSOC}, as a similar endeavor, seeks to enhance classification performance by distinguishing between classification targets and background elements. 
HCTransformers~\cite{HCTransformers} propose a hierarchical cascading transformer architecture, aiming to address the overfitting challenges faced by large-scale models in few-shot learning. 
Meanwhile, FewTURE~\cite{FewTURE} similarly employs transformer architecture to extract key features from the main subjects within images. 
In the realm of generalized few-shot learning, a substantial body of work~\cite{CoOp,TipAdapter} has already leveraged pre-trained models to enhance the efficacy of few-shot learning. 
In our research, we have also incorporated the pre-trained CLIP~\cite{CLIP} model to enhance the feature extraction capabilities of our model. The critical distinction, however, lies in the fact that our model is trained using a meta-learning approach. 

\subsection{Sample Relation}
There exist diverse sample relationships among different class samples, and currently, most models are built upon the foundation of establishing these sample relationships. Numerous studies aim for models to achieve strong generalization performance across various class sample relationships, thereby minimizing vicinal risk. 
CAN~\cite{CrossAttNet} and OLTR~\cite{OLTR} incorporate sample-specific relationships within the shared context by leveraging the correlations among individual samples. 
IEM~\cite{IEM} analyzes local correlations among samples and performs memory storage updates for these correlations. 
IRM~\cite{IRM} achieves a reduced vicinal risk by exploring the correlation between sample invariant features and spurious features. 
In cross-domain tasks, ~\cite{cdg} explores the transferability of sample relationships across different domains by discarding specific sample relationships. 
Similar to ~\cite{cdg}, ~\cite{DAL} explores domain-invariant and class-invariant relationships by employing the deep adversarial disentangled autoencoder to achieve cross-domain classification tasks. 
BatchFormer~\cite{batchformer} has achieved significant improvements across various data scarcity tasks by implicitly exploring the relationships among mini-batch samples during training. 
In mixup~\cite{mixup}, samples are linearly interpolated to capture the class-invariant relationships between samples. 
In our work, we perform linear combinations of samples to explore task-relevant relationships among them. 
\subsection{Contrative Learning}
Contrastive learning has achieved significant success in recent years. 
InstDisc~\cite{instdisc} proposes the utilization of instance discrimination tasks as an alternative to class-based discrimination tasks within the framework of unsupervised learning. 
MOCO~\cite{MOCO} achieves favorable transferability to downstream tasks through the strategy of constructing a dynamic dictionary and performing momentum-based updates. 
Contrastive learning has exhibited its generality and flexibility in time series tasks, encompassing domains like audio and textual data. 
An abundance of work~\cite{MOCO,SimClr,CLIP} has demonstrated the positive impact of contrastive learning in both unsupervised learning and generalization research within the realm of computer vision. 
The objective of contrastive learning is to bring together samples of the same class while separating those from different classes, thus constructing suitable patterns for sample feature extraction. 
In episodic training, we utilize contrastive learning methods to extract class relationships within the task, enhancing the classification performance for few-shot learning.
\section{Method}
In this section, we first describe the problem definition related to few-shot learning. Subsequently, an exposition of our proposed methodology is presented. Conclusively, we delve into a comprehensive discussion on the two important components of our method: Prototype Extraction Module and Prototype Contrastive Loss.

\subsection{Problem Formulation}
Episodic training differs from the deep neural networks training approach. 
In the traditional training of deep neural networks, we usually train the neural network on a sample-by-sample basis. 
In episodic training, we typically train the neural network on a task-by-task basis. 
The episodic training mechanism~\cite{matchnet} has been demonstrated to facilitate the learning of transferable knowledge across classes. 

In few-shot learning, we usually divide the dataset into training, validation, and test sets. 
The training set, validation set, and test set have no overlapping classes. Therefore, we refer to the classes in the training set as seen classes, while the classes in the validation set and test set are termed unseen classes. 
During the training phase, we randomly sample from the training set to create the support set and the query set. 
We use $S$ to represent the support set and $Q$ to define the query set. 
In the support set $S$, there are $N$ classes, and each class contains $K$ samples. 
We treat the query set $Q$ as unlabeled samples and perform classification on the unlabeled samples in $Q$ using the labeled samples in the support set $S$, which contains $N$ classes, each with $K$ samples. 
During the testing phase, we follow the same procedure and divide the test set into a support set and a query set, similar to what we did during the training phase. 
This allows us to evaluate the few-shot learning performance of the model on unseen classes in a manner consistent with the training process. 
We typically refer to tasks that satisfy the above settings as N-way K-shot tasks. 
In our work, we train and evaluate the model using the aforementioned problem formulation. 
  
\begin{figure}[!t]
\centering
\includegraphics[scale=0.9]{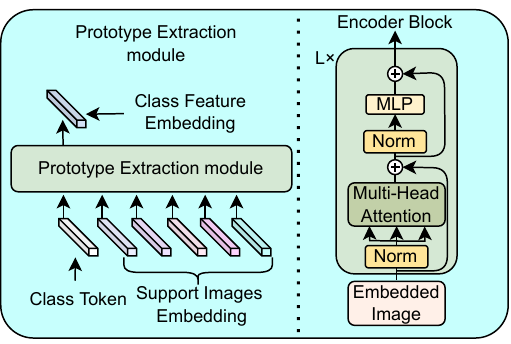}
\caption{The prototype extraction module adopts the transformer structure~\cite{Transformer}, taking the prototype token and embeddings of same-class images from the support set as inputs to obtain the prototype and sub-prototype for that class.}
\label{cls_module}
\end{figure}

\subsection{Overview}
We linearly combine the support set and apply non-linear mapping through the prototype extraction module. Furthermore, we optimize the prototype extraction module using contrastive learning strategies to attain improved prototype representations. As illustrated in Figure~\ref{overview}, we process both the support set and query samples through a frozen CLIP feature extraction network to obtain image embeddings. 
Subsequently, we perform linear combinations on the support set samples to generate $C^1_K$ sub-support sets. 
Simultaneously, a prototype token is added to each support set and sub-support set, derived by computing the average of the respective embedding collection. 
Individually, each support set and sub-support set is fed into the prototype extraction module to obtain encoded prototypes and sub-prototypes. 
We retain the prototypes and sub-prototypes while discarding the sample embeddings from the support sets. 
We compute $L_{Prototype}$ using the retained sub-prototypes through contrastive loss, while $L_{Classifier}$ is obtained by calculating the embeddings of query samples and prototypes. 
Finally, we sum up $L_{Prototype}$ and $L_{Classifier}$ to create the ultimate optimization objective. 

\subsection{Prototype Extraction Module }
In this section, we will provide a comprehensive exposition of our proposed prototype extraction module. 
Additionally, we will conduct a comparative analysis between our method and existing class feature extraction approaches found in the paper.

First we introduce the prototype representation, the earliest class feature representation to appear in few-shot learning. 
In the N-way K-shot task, we assume the existence of a class C, and in the support set $S$, there exists a subset $S_C=\left \{ x_1,x_2,\dots x_K \mid y=C  \right \} $. 
We refer to the feature extraction network as $f$. 
In that case, we can express the class feature representation in the prototypical networks~\cite{protonet} as follows:
\begin{equation}
    Prototype\left ( C \right ) =\frac{1}{K} \sum_{i=1}^{K}f\left ( x_i \right )  ,  x_i\subset S_C
\end{equation}
The method of prototype points provides a simple and effective way to express class features. 
Absolutely, the global average pooling layer used in the feature extraction network can introduce noise into the prototype points, causing them to deviate from their true representation and leading to bias. 
To address this issue, DN4~\cite{DN4} and DMN4~\cite{DMN4} remove the global average pooling layer from the feature extraction network. 
They employ local descriptors to replace the global feature representation of images and utilize a discriminative nearest neighbor algorithm to obtain the most representative local descriptors in the images as the feature representation for samples. 

However, we believe that the image background has a certain influence on the image classification performance and also provides some category-related contextual features. 
Therefore, we propose a novel class feature extraction module referred to as prototype extraction module to replace the current few-shot class feature representation. 
In ViT~\cite{VIT}, the image is divided into patches, and transformer~\cite{Transformer} is utilized to compute the correlations between these patches, resulting in the overall feature representation of the entire image. 
Inspired by ViT, we simply treat the image as a set of patches input to the transformer, thereby obtaining the feature representation for the entire class. 
The fundamental architecture of prototype extraction module is illustrated in Figure \ref{cls_module}. 
We use $\phi $ to represent the prototype extraction module, and we can express it in the following form:
\begin{equation}
Prototype\left ( C \right ) =
\phi \left (x_{token}, f\left ( x_1 \right ),f\left ( x_2 \right ),\dots f\left ( x_K \right )   \right ),x_i\subset S_C
\end{equation}
In the formula, $x_{token}$ represents the prototype token for that class, and it can be expressed as:
\begin{equation}
x_{token}=\frac{1}{K}\sum_{i=1}^{K}f\left ( x_i \right ) ,x_i\subset S_C
\end{equation}
Finally, we use a simple metric learning classification method to classify the query samples. 
Specifically, we calculate the distance between the embeddings of the query samples and the prototype points in the feature space to measure the similarity between the query samples and each class. 
This distance metric is used for classification, where the query sample is assigned to the class with the closest feature embedding in the feature space. 
This classification approach can be formalized with the following formula: 
\begin{equation} 
argmin_{ c\subset C } L_2\left ( x_{query},Prototype\left ( c \right )  \right ) 
\end{equation}
The classification loss is optimized using the cross-entropy loss, and the formula for the classification loss is as follows: 
\begin{equation}
   Loss_{classify}=
-\sum_{c=1}^{N}y_{c}log\left ( \frac{e^{-L_2\left ( x_{query}, Prototype \left ( c \right )  \right )} }{\sum_{i=1}^{N}e^{-L_2\left ( x_{query}, Prototype \left ( i \right )  \right )} }\right )
\end{equation}
The $y_{c}$ is the one-hot encoding of the true class label for the sample. 
\subsection{Prototype Contrastive Loss}
To enhance the generalization capability of the prototype extraction module, we drew inspiration from contrastive learning and proposed prototype contrastive loss. 
The contrastive loss was first introduced by ~\cite{contloss} and laid the foundation for subsequent highly successful contrastive learning~\cite{MOCO,SimClr}. 
The main idea of the contrastive loss is to construct positive and negative sample pairs, where positive pairs are brought closer together in the feature space, while negative pairs are pushed further apart. 

In few-shot learning, by extracting K-1 samples from the same class in the support set $S$, we can obtain $K$ different sub-support set of samples $S_{ci}=\left \{ x_{c1},\dots ,x_{ci-1},x_{ci+1}\dots ,x_{cK} \right \} ,i=1,2\dots K,c\subset C$. 
Then, we pass each of these K sub-support sets constructed from the same class samples through the prototype extraction module to obtain K sub-prototypes for that class. 
We use the K sub-prototypes obtained from the same-class support set samples as positive pairs. 
At the same time, we use the sub-prototypes obtained from different-class sub-support sets as negative pairs. 
We represent the constructed positive sample pairs as follows:
\begin{equation}
    Pos_c=\left \{ p_{c1},p_{c2},\dots p_{cK} \right \} ,C=1,2\dots N
\end{equation}
Thus, we can obtain the prototype contrastive loss using the constructed positive and negative pairs as follows: 
\begin{equation}
   L_{prototype}=exp\left ( \frac{1}{N} \cdot \frac{\sum_{i,j=1}^{K} L_2\left ( p_{ci},p_{cj} \right )+I}{\sum_{m\ne n}^{}\sum_{i,j=1}^{K}  L_2\left ( p_{m,i},p_{n,j} \right )+I}  \right ) 
\end{equation}

Because when $K=1$, the support set contains only one sample per class, leading to $\sum_{i,j=1}^{K} L_2\left ( p_{ci},p_{cj} \right )=0$. To avoid this situation, we add the identity element I to prevent it from happening. 
The overall loss of the model during the training phase is as follows:
\begin{equation}
    Loss=Loss_{classifier}+Loss_{prototype}
\end{equation}
Finally, we present the pseudocode for the training process of PrototypeFormer in Algorithm~\ref{prototypeformer}.

\begin{algorithm}[h]
\caption{Training Process of PrototypeFormer}
\label{prototypeformer}
\begin{algorithmic}[1]
\State \textbf{Input:}
\State \quad Support set $S = \{S_1, S_2, \dots, S_N\}$, where $S_c = \{x_{c1}, x_{c2}, \dots, x_{cK}\}$ for class $c$. Query set $Q = \{x_{q1}, x_{q2}, \dots, x_{qM}\}$. Pre-trained CLIP feature extractor $f$ (frozen). Prototype extraction module $\phi$ (Transformer-based). Number of classes $N$, number of shots $K$.
\State \textbf{Output:}
\State \quad Classification results for query set $Q$.

\State \textbf{Step 1: Extract features for support and query sets}
\For{each $x \in S \cup Q$}
    \State $x_{\text{embedding}}, x_{q_{\text{embedding}}} = f(x)$ \Comment{Extract features using CLIP}
\EndFor

\State \textbf{Step 2: Generate sub-support sets and sub-prototypes}
\For{each class $c = 1, 2, \dots, N$}
    \State $S_c = \{x_{c1}, x_{c2}, \dots, x_{cK}\}$ \Comment{Support set for class $c$}
    \State $\text{sub\_S}_c = \text{generate\_sub\_support\_sets}(S_c, K)$ \Comment{Generate $K$ sub-support sets}
    \For{each sub-support set $\text{sub\_set} \in \text{sub\_S}_c$}
        \State $x_{\text{token}} = \frac{1}{K-1} \sum_{x_i \in \text{sub\_set}} f(x_i)$ \Comment{Compute prototype token}
        \State $\text{sub\_prototype} = \phi(x_{\text{token}}, \text{sub\_set})$ \Comment{Extract sub-prototype}
        \State $\text{sub\_prototypes}_c.\text{append}(\text{sub\_prototype})$ \Comment{Store sub-prototype}
    \EndFor
\EndFor

\State \textbf{Step 3: Compute prototype contrastive loss}
\State $L_{\text{prototype}} = 0$
\For{each class $c = 1, 2, \dots, N$}
    \State Positive pairs, Negative pairs $\leftarrow$ sub-prototypes of the same class, sub-prototypes of different classes
    \State $L_{\text{prototype}} += \text{contrastive\_loss}(\text{pos\_pairs}, \text{neg\_pairs})$ \Comment{Compute contrastive loss}
\EndFor

\State \textbf{Step 4: Compute classification loss}
\State $\text{prototypes} = \left\{\frac{1}{K} \sum_{p \in \text{sub\_prototypes}_c} p \mid c = 1, 2, \dots, N\right\}$ \Comment{Compute prototypes}
\State $L_{\text{classifier}} = 0$
\For{each query sample $x_q \in Q$}
    \State $\text{distances} = \left\{L_2(x_{q_{\text{embedding}}}, \text{prototypes}_c) \mid c = 1, 2, \dots, N\right\}$ \Comment{Compute distances}
    \State $L_{\text{classifier}} += \text{cross\_entropy\_loss}(\text{distances}, \text{true\_label})$ \Comment{Compute classification loss}
\EndFor

\State \textbf{Step 5: Optimize the model}
\State $\text{Loss} = L_{\text{classifier}} + L_{\text{prototype}}$
\end{algorithmic}
\end{algorithm}

\begin{table*}[t]
\caption{Few-shot learning classification accuracies(\%) on miniImageNet, tieredImagenet and CUB-200 under the setting of 5-way 1-shot and 5-way 5-shot with 95\% confidence interval. (`-' not reported)}
\label{experimenttable}
\resizebox{\linewidth}{!}{
\renewcommand\arraystretch{1.2}
\begin{tabular}{lcccccc}
\hline
\multicolumn{1}{c}{\multirow{2}{*}{Model}} &
  \multicolumn{2}{c}{miniImageNet} &
  \multicolumn{2}{c}{tieredImagenet} &
  \multicolumn{2}{c}{CUB-200} \\ \cmidrule(r){2-3} \cmidrule(r){4-5} \cmidrule(r){6-7}
\multicolumn{1}{c}{} & 5-way 5-shot   & 5-way 1-shot   & 5-way 5-shot   & 5-way 1-shot          & 5-way 5-shot   & 5-way 1-shot        \\ \hline
MAML~\cite{MAML}                 & 64.31±1.1    & 47.78±1.75   & 71.10±1.67   & 52.07±0.91          &     -           &      -               \\
Prototypical Network~\cite{protonet} & 78.44±0.21   & 60.76±0.39   & 80.11±0.91   & 66.25±0.34          &       -          &          -            \\
HCTransformers~\cite{HCTransformers}       & 89.19 ± 0.13 & 74.62 ± 0.20 & 91.72 ± 0.11 & 79.57 ± 0.20        &       -          &         -             \\
DeepEMD~\cite{Deepemd}              & 82.41 ± 0.56 & 65.91 ± 0.82 & 86.03 ± 0.58 & 71.16 ± 0.87        & 88.69 ± 0.50 & 75.65 ± 0.83      \\
MCL~\cite{MCL}                  & 83.99        & 67.51        & 86.02        & 72.01               & 93.18        & 85.63             \\
POODLE~\cite{POODLE}               & 85.81        & 77.56        & 86.96        & 79.67               & 93.80        & 89.88             \\
FRN~\cite{FRN}                  & 82.83±0.13   & 66.45±0.19   & 86.89±0.14   & 72.06±0.22          & 92.92±0.10   & 83.55±0.19        \\
PTN~\cite{PTN}                  & 88.43±0.67   & 82.66±0.97   & 89.14±0.71   & 84.70±1.14          &        -         &           -           \\
FewTURE~\cite{FewTURE}              & 86.38±0.49   & 72.40±0.78   & 89.96±0.55   & 76.32±0.87          &         -        &            -          \\
EASY~\cite{EASY}                 & 89.14 ± 0.1  & 84.04 ± 0.2  & 89.76 ± 0.14 & 84.29 ± 0.24        & 93.79 ± 0.10 & 90.56 ± 0.19      \\
 iLPC~\cite{iLPC}                 & 88.82±0.42   & 83.05±0.79   & 92.46±0.42   &\textbf{ 88.50±0.75} & 94.11±0.30     & \textbf{91.03±0.63} \\
Simple CNAPS~\cite{SimpleCNAPS}         & 89.80        & 82.16        & 89.01        & 78.29               &       -          &            -          \\
MBSS~\cite{mbss} & 86.32 ± 0.44 & 78.93 ± 0.82 & 91.41 ± 0.48 & 87.42 ± 0.82 & 90.83±0.39 & 86.26±0.74 \\ 
BRAVE~\cite{ji2024brave}&88.93 ± 0.32 &68.55 ± 0.28 &89.05 ± 0.24 &73.79 ± 0.44 &-&-\\
   FGFD GNN~\cite{ganesan2024few}&96.50 ± 0.25 &81.65 ± 0.98 & -&- & 91.56 ± 0.24 &78.93 ± 0.42 \\
FeatWalk~\cite{chen2024featwalk}&87.38 ± 0.27 &70.21 ± 0.44&  89.92 ± 0.29 &75.25 ± 0.48 &\textbf{95.44 ± 0.16} &85.67 ± 0.38 \\
  
\hline
Ours &
  \textbf{97.07 ± 0.11} &
  \textbf{90.88 ± 0.31} &
  \textbf{95.00 ± 0.19} &
  87.26 ± 0.40 &
  94.25 ± 0.16 &
  89.04 ± 0.35 \\ \hline
\end{tabular}
}

\end{table*}

\section{Experiments}

In this section, we will evaluate the proposed method on multiple few-shot benchmark datasets and compare it with state-of-the-art methods. 
Additionally, we will conduct ablation experiments and visualization experiments to further analyze and validate the effectiveness of our proposed approach. 

\subsection{Datasets}

\textbf{miniImageNet}~\cite{matchnet} is a subset of the larger ImageNet dataset and is widely used in few-shot learning research. 
It consists of 100 classes, with each class containing 600 images, resulting in a total of 60,000 images. 
The dataset is divided into 64 classes for the training set, 16 classes for the validation set, and 20 classes for the test set. 

\textbf{tieredImagenet} is a larger subset of the ImageNet dataset compared to miniImagenet. 
The dataset consists of 608 classes with a total of 779,165 images. 
For few-shot learning, it is divided into three subsets, with 351 classes used for the training set, 97 classes for the validation set, and 160 classes for the testing set. 

\textbf{Caltech-UCSD Birds-200-2011}~\cite{CUB}, also known as CUB, is the benchmark image dataset for current fine-grained classification and recognition research. 
The dataset contains 11,788 bird images, encompassing 200 subclasses of bird species. 
We split it into 100, 50, and 50 classes for training, validation, and testing, respectively. 

\subsection{Experimental Settings}
To obtain better image features, we use ViT-Large/14 as the backbone for image feature extraction and pair it with the same CLIP pre-trained model used in CoOp~\cite{CoOp} and Clip-Adapter~\cite{ClipAdapter}. 
Due to the limited data in the context of few-shot learning, prototype extraction module adopts a two-layer transformer architecture without incorporating positional encoding. 
During the training phase, we freeze the feature extraction network and only train the prototype extraction module proposed in this paper to preserve the image feature extraction capabilities of the pre-trained CLIP model and obtain a prototype extraction module with excellent class feature representations. 
  
During the training phase, we maintain the traditional episodic training approach and conduct training on 5-way 5-shot and 5-way 1-shot task settings. 
Additionally, we use the Adam~\cite{ADAM} optimizer to optimize the model. 
We set the initial learning rate of the optimizer to 0.0001. 
The momentum weight coefficients $\beta$1 and $\beta$2, as well as the $\epsilon$ parameter of the optimizer, are set to their default values of 0.9, 0.999, and 1e-8, respectively. 
In the gradient updating strategy, we adopt the gradient accumulation algorithm, where we accumulate gradients over every 10 batches before performing a parameter update. 
We train the model for 100 epochs, where each epoch consisted of 500 batches, and each batch represented a task. 
In image augmentation, we resize the images and then apply center cropping to obtain $224 \times 224$ pixel image inputs. 

In the testing phase, to ensure fairness, we adhere to the evaluation methodology of few-shot learning without making any changes. 
We randomly sample 2000 tasks from the test set. For each task, we extract 15 query samples per class to evaluate our method. 
We report the average accuracy with a 95\% confidence interval to ensure the reliability of our results. 

\begin{figure*}[!t]
\centering
\includegraphics[width=1\linewidth]{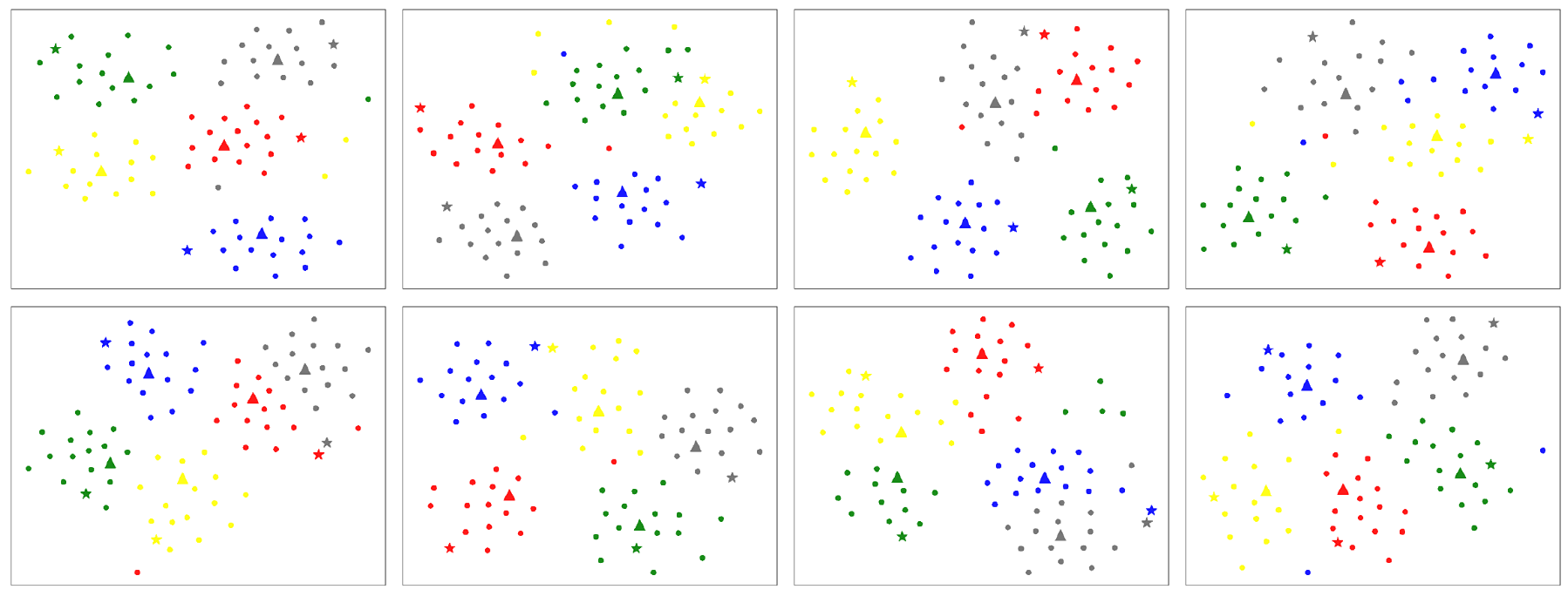}
\caption{We randomly select eight task sets from the test dataset and visualize their feature embeddings using t-SNE~\cite{TSNE}. In the visualization, circular points represent query samples, triangles represent prototype points obtained by averaging the support set, and pentagrams represent class feature embeddings obtained through our proposed method in this paper.}
\label{scatter}
\end{figure*}

\subsection{Results}
Following the few-shot standard experimental settings, we conduct experiments on both 5-way 1-shot and 5-way 5-shot tasks to evaluate our method. 
The experimental results are presented in Table \ref{experimenttable}. 

As shown in the table \ref{experimenttable}, our method outperforms the current state-of-the-art results on both 5-way 5-shot and 5-way 1-shot tasks in the miniImageNet dataset. 
Excitingly, our method achieve an accuracy improvement of 0.57\% over the current state-of-the-art method in the 5-way 5-shot task on this dataset. 
At the same time, our method also achieve a 6.84\% accuracy improvement in the 5-way 1-shot task compared to the current state-of-the-art method. 
Our method achieve significant improvements in the 5-way 5-shot task on both the tieredImagenet dataset and the CUB-200 dataset compared to the existing methods. 
Observing the table, we can notice that compared to the 5-way 5-shot tasks, our method's performance is slightly inferior in the 5-way 1-shot tasks. 
We believe that this is due to the lack of positive pairs in the 5-way 1-shot task, which hinders the prototype extraction module's ability to represent class features accurately. 

\subsection{Ablation Study}

To validate the effectiveness of our method, we conduct ablation experiments from various perspectives on the proposed approach. 
 
\begin{table*}[!t]

\caption{This ablation experiment aims to validate the effectiveness of the prototype extraction module.}
\renewcommand\arraystretch{1.2}
\begin{tabular}{ccccccc}
\hline
\multirow{2}{*}{Model} & \multicolumn{2}{c}{miniImageNet}  & \multicolumn{2}{c}{tieredImagenet} & \multicolumn{2}{c}{CUB-200}       \\ \cmidrule(r){2-3} \cmidrule(r){4-5} \cmidrule(r){6-7}
     & 5-way 5-shot    & 5-way 1-shot    & 5-way 5-shot    & 5-way 1-shot    & 5-way 5-shot    & 5-way 1-shot    \\ \hline
CLIP & 95.13 ± 0.14\% & 83.86 ± 0.40\% & 92.25 ± 0.24\% & 79.24 ± 0.46\% & 89.20 ± 0.24\% & 72.51 ± 0.51\% \\
Ours & 97.07 ± 0.11\% & 90.88 ± 0.31\% & 95.00 ± 0.19\% & 87.26 ± 0.40\% & 94.25 ± 0.16\% & 89.04 ± 0.35\% \\ \hline
\end{tabular}

\label{ablation1}
\end{table*}

To validate the effectiveness of prototype extraction module, we conduct ablation experiments under two conditions: removing the prototype extraction module and retaining the prototype extraction module as part of our method. 
The experimental results are shown in Table \ref{ablation1}, where ``CLIP" represents the condition where we remove the prototype extraction module and retain only the CLIP pre-trained model. 
From the Table \ref{ablation1}, we can observe that the CLIP pre-trained model itself exhibits good few-shot image classification performance due to its strong zero-shot knowledge transfer ability in few-shot learning. 
Furthermore, our proposed method shows significant performance improvement compared to the comparative methods in the ablation experiments. 

As shown in Table \ref{ablationloss}, we conduct experiments on the miniImageNet dataset in both 5-way 5-shot and 5-way 1-shot settings with and without the inclusion of the prototype contrastive loss. 
The experimental results indicate that the prototype loss has a positive impact on model optimization. 
Additionally, in Table \ref{Lblock}, we conduct ablation experiments on prototype extraction modules with 2, 4 and 6 layers of transformer blocks. 

\begin{table}[!t]
\caption{The table presents a comparative experiment on whether to include the prototype contrastive loss in the model.}
\renewcommand\arraystretch{1.2}
\begin{tabular}{cll}
\hline
\multirow{2}{*}{Model} & \multicolumn{2}{c}{miniImageNet}                                    \\ \cline{2-3} 
                       & \multicolumn{1}{c}{5-Way 5-Shot} & \multicolumn{1}{c}{5-Way 1-Shot} \\ \hline
L\_classifier            & 96.24 ± 0.11\%                  & 89.13 ± 0.32\%                  \\
L\_classifier+L\_prototype     & 97.07 ± 0.11\%                  & 90.88 ± 0.31\%                  \\ \hline
\end{tabular}

\label{ablationloss}
\end{table}

\begin{figure}[!t]
\centering
\includegraphics[width=0.85\linewidth]{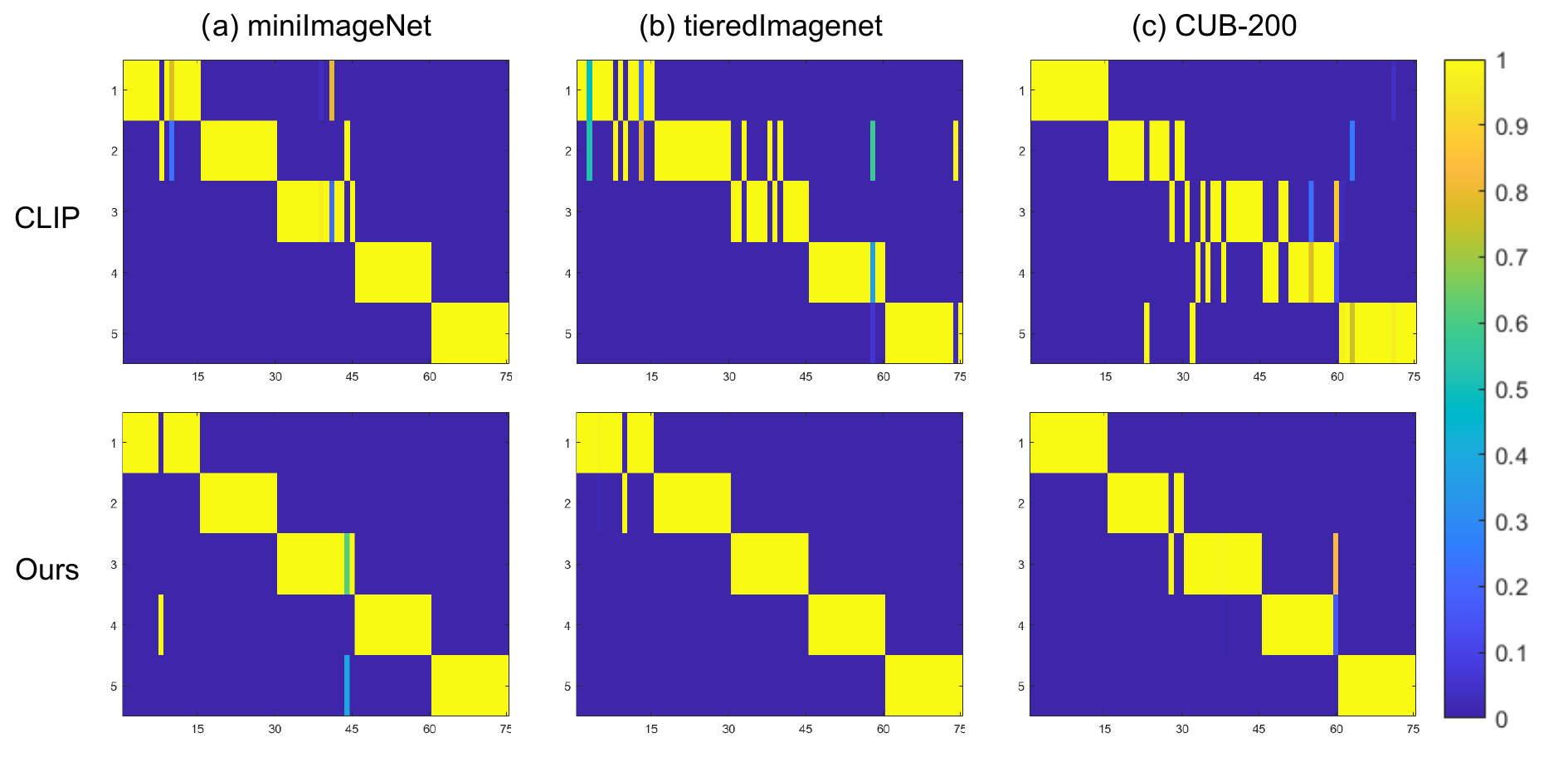}
\caption{We randomly choose 5 categories from the test set, with 15 samples in each category, and create their similarity matrix. In the visualization, yellow areas show correct classifications, while blue areas indicate misclassifications.}

\label{heatmap}
\end{figure}

\subsection{Visualization}

In this section, we delve into a comprehensive visualization analysis based on the model trained on the 5-way 5-shot task of the miniImageNet dataset. The visualization, depicted in Figure \ref{scatter}, involves the random extraction of samples from 8 tasks in the test set, showcasing them using t-SNE. The visualization emphasizes the 15 query set samples through circular symbols, while triangular symbols signify the prototype points derived by averaging the embeddings of support set samples. Additionally, pentagram symbols denote the prototypes obtained using the prototype extraction module introduced in this paper.

Upon careful observation of Figure \ref{scatter}, a notable distinction emerges. Class embeddings obtained through the prototype point calculation method, as seen in prototypical networks~\cite{protonet}, tend to be positioned relatively closer to the center of their respective classes. In contrast, the class embeddings derived from our proposed method are strategically positioned towards the edges of the respective classes. This distinction arises from the underlying objectives of the two methods. The prototype point calculation method aims to represent the inherent characteristics of each class, positioning prototype points at the center to describe the class distribution in the feature space. On the other hand, our method strategically places class embeddings towards the edges, aiming to maximize the separation from other class samples while staying close to samples of the same class for effective classification.

To further underscore the efficacy of our approach, we conduct a matrix similarity visualization comparing our method with the traditional prototype point approach, as illustrated in Figure \ref{heatmap}. Notably, the term "CLIP" refers to the conventional prototype point representation using the CLIP pre-trained model as the backbone. These experiments are conducted separately on the miniImageNet, tieredImagenet, and CUB-200 datasets. The results showcased in Figure \ref{heatmap} unequivocally highlight the substantial enhancement achieved by our proposed method in the domain of few-shot classification.

\begin{table}[t]
\caption{Ablation experiments of prototype extraction module with 2, 4 and 6 transformer blocks on miniImageNet dataset.}
\begin{center}
\begin{tabular}{p{2.2cm}cc}
\hline
\centering
\renewcommand\arraystretch{1.8}
\multirow{2}{*}{L$\times$block} & \multicolumn{2}{c}{miniImageNet}  \\ \cline{2-3} 
                         & 5-way 5-shot    & 5-way 1-shot    \\ \hline
2                        & 97.07 ± 0.11\% & 90.88 ± 0.31\% \\
4                        & 95.96 ± 0.13\% & 90.03 ± 0.33\% \\
6                        & 94.44 ± 0.17\% & 88.33 ± 0.35\% \\\hline
\end{tabular}
\end{center}
\label{Lblock}
\end{table}
 
\section{Conclusions}

We propose PrototypeFormer, a simple transformer-based backbone for exploring the relationships among prototypes of few-shot classes to enhance the capability of robust feature learning. To further enhance the discriminative characteristics of prototype features, we introduce prototype contrastive learning for the optimization of prototypes. In contrast to instance discrimination, we treat sub-prototypes from the same category as positive samples and sub-prototypes from different categories as negative samples. We evaluate PrototypeFormer on several popular few-shot image classification benchmark datasets and conduct comprehensive analyses through ablation experiments and visualization techniques. The experimental results demonstrate that our approach significantly outperforms the current state-of-the-art methods. The success of PrototypeFormer is further evidenced by its ability to generalize well across diverse datasets, showcasing its robustness and versatility in various image classification challenges. We hope that our work encourages further exploration into sample relations, prototype relations, and class relations in few-shot learning.

\printcredits
 
\noindent \textbf{Declaration of competing interest} 

The authors declare that they have no known competing financial
 interests or personal relationships that could have appeared to
influence the work reported in this paper.

\noindent \textbf{Acknowledgements}

This work was supported in part by the Priority Academic Program Development of Jiangsu Higher Education Institutions, and by the National Natural Science Foundation of China under Grant No.61672364, No.62176172 and No.61902269.

\noindent \textbf{Data Availability }

The datasets used during this study are available upon reasonable request to the authors.



\bibliographystyle{unsrt}
\bibliography{cas-refs}
\nocite{*}

\end{document}